\documentclass[a4paper,twoside]{article}

\usepackage{epsfig}
\usepackage{subcaption}
\usepackage{calc}
\usepackage{amssymb}
\usepackage{amstext}
\usepackage{amsmath}
\usepackage{amsthm}
\usepackage{multicol}
\usepackage{multirow}
\usepackage{pslatex}
\usepackage{apalike}
\usepackage{SCITEPRESS}
\usepackage{comment}

\begin{document}

\title{Performance Analysis of a Foreground Segmentation Neural Network Model}

\author{\authorname{Joel Tom\'{a}s Morais\sup{1}, Ant\'{o}nio José Borba Ramires Fernandes\sup{1}, Andr\'{e} Leite Ferreira\sup{2} and Bruno Faria\sup{2}}
\affiliation{\sup{1}Minho University}
\affiliation{\sup{2}Bosch Car Multimedia Portugal S.A.}
\email{ a70841@alunos.uminho.pt, arf@di.uminho.pt,\{Andre.Ferreira2, Bruno.Faria\}@pt.bosch.com}
}


\abstract{In recent years the interest in segmentation has been growing, being used in a wide range of applications such as fraud detection, anomaly detection in public health and intrusion detection. 
We present an ablation study of FgSegNet\_v2, analysing its three stages: (i) Encoder, (ii) Feature Pooling Module and (iii) Decoder. The result of this study is a proposal of a variation of the aforementioned method that surpasses state of the art results.
Three datasets are used for testing: CDNet2014, SBI2015 and CityScapes. In CDNet2014 we got an overall improvement compared to the state of the art, mainly in the LowFrameRate subset. The presented approach is promising as it produces comparable results with the state of the art (SBI2015 and Cityscapes datasets)  in very different conditions, such as different lighting conditions}

\onecolumn \maketitle \normalsize \setcounter{footnote}{0} \vfill


\section{\uppercase{Introduction}}
\label{sec:introduction}

\noindent

\noindent Over the years object detection has seen different application domains with the aim of detecting an object type and location in a specific context of an image. Nevertheless, for some applications detecting an object location and type is not enough. For these cases, the labelling of each pixel according to its surroundings in an image presents itself as an alternative approach. This task is known as segmentation~\cite{Hafiz2020} and finds applications in video monitoring, intelligent transportation, sports video analysis, industrial vision, amongst many other fields~\cite{Setitra2014}. Some of the techniques used in traditional segmentation include Thresholding, K-means clustering, Histogram-based image segmentation and Edge detection. Throughout the last years, modern segmentation techniques have been powered by deep learning technology. 

 
Segmentation methods isolate the objects of interest of a scene by segmenting it into background and foreground. Throughout the aforementioned applications, segmentation is usually applied to localizing the objects of interest in a video using a fixed camera, presenting robust results.

Nevertheless, when the camera is not fixed or the background is changing, detecting objects of interest using background subtraction could be more demanding, generating diverse false positives. Moreover, it demands a high computational cost when applied to videos~\cite{Liu2020}. 
Another issue with the method consists in developing a self-adaptive background environment, accurately describing the background information. This can be challenging since the background could be changing a lot, e.g., in lighting and blurriness~\cite{Minaee2020}.

After exploring numerous state of the art methods in this field, FgSegNet\_v2 was chosen as a suitable candidate to explore since it outperforms every state of the art method in the \cite{ChangeDetection1} challenge. 
Through the analysis of its components, and exploring variations for each, we have achieved a more robust method that can cope with both fixed and moving cameras amongst datasets with different scenarios.

This work is organized as follows: section \ref{sec:related} describes the current state of the art in instance segmentation; section \ref{sec:exploring} presents FgSegNet\_v2 architecture and our proposed variations; in section \ref{sec:results} we introduce the used metrics and datasets, followed by the experiments made in the FgSegNet\_v2, followed by our results showing that the current implementation achieves state of the art performance. A conclusion and some avenues for future work are presented in section \ref{sec:conclusion}.


\section{\uppercase{Related Work}}
\label{sec:related}

\noindent Background subtraction has been studied in the Statistics field since the 19th century~\cite{Chandola2009}. Nowadays, this subject is being strongly developed in the Computer Science field, from using arbitrary to automated techniques~\cite{Lindgreen2004}, in Machine Learning, Statistics, Data Mining and Information Theory. 

Some of the earlier background subtraction techniques are non-recursive, such as \emph{Frame differencing}, \emph{Median Filter} and \emph{Linear predictive filter}, in which a sliding window approach is used for the background estimation~\cite{Sen-Ching2004}, maintaining a buffer with past video frames and estimating a background model based on the statistical properties of these frames, resulting in a high consumption of memory~\cite{Parks2008}.
As for the recursive techniques, \emph{Approximated Median Filter}, \emph{Kalman filter} and \emph{Gaussian related} do not keep a buffer for background subtraction, updating a single background based on the input frame~\cite{Sen-Ching2004}. Due to the recursive usage, by maintaining a single background model which is being updated with each new video frame, less memory is going to be used when compared to the non-recursive methods~\cite{Parks2008}.

Regarding segmentation, and using DL, the top state of the art techniques include Cascade CNN~\cite{Wang2017-2}, \emph{FgSegNet\_v2}~\cite{Lim2019} and \emph{BSPVGAN}~\cite{Zheng2020}. 
These methods have achieved the top three highest scores in the \cite{ChangeDetection1} challenge.

\emph{Cascade CNN}~\cite{Wang2017-2} is a semi-automatic method for segmenting foreground moving objects, it is an end-to-end model based on Convolutional Neural networks (CNN) with a cascaded architecture. This approach starts by manually selecting a sub-set of frames containing objects properly delimited, which are then used to train a model. The model embodies three networks, differentiating instances, estimating masks and categorizing objects. These networks form a cascaded structure, allowing them to share their convolutional features while also being able to generalize to cascades that have more stages, maintaining the prediction phase extremely fast.

In \emph{FgSegNet\_v2}~\cite{Lim2019}, the aim is to segment moving objects from the background in a robust way under various challenging scenarios. It can extract multi-scale features within images, resulting in a sturdy feature pooling against camera motion, alleviating the need of multi-scale inputs to the network. A modified VGG 16 is used as an encoder for the network, obtaining higher-resolution feature maps, which will be used as input for the Feature Pooling Module (FPM) and consequently as input for the decoder, working with two Global Average Pooling (GAP) modules.

\emph{BSPVGAN}~\cite{Zheng2020} starts by using the median filtering algorithm in order to extract the background. Then, in order to classify the pixels into foreground and background, a background subtraction model is built by using Bayesian GANs. Last, parallel vision~\cite{Wang2017} theory is used in order to improve the background subtraction results in complex scenes.  
Even though the overall results in this method don't outperform the FgSegNet\_v2, it shows some improvements regarding lighting changes, outperforming other methods in Specificity and False Negative changes.


\section{\uppercase{Exploring Variations for FgSegNet\_v2}}
\label{sec:exploring}

\noindent In this section, an ablation study of the FgSegNet\_v2 components is presented, in order to potentially improve the previous method by reworking individually these components, obtaining a more robust method.
The section is divided in different modules, starting from the details of the architecture, along with the description of the elements in it, i.e, Encoder, Feature Pooling module and Decoder, ending with the training details.\\

\begin{figure*}
\begin{center}
\includegraphics[scale=0.6]{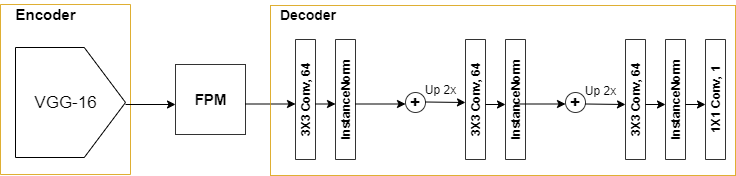}
\end{center}
   \caption{Proposed method architecture.}
\label{fig:arch}
\end{figure*}

Following a segmentation strategy, the proposed work intends to segregate the background pixels from the foreground ones. In order to achieve this, a VGG-16 architecture is used as an encoder, in which the corresponding output is used as input to a FPM ( Feature Pooling Module ) and consequently to a decoder. The overall architecture can be seen in Figure \ref{fig:arch}.\\

\subsection{Encoder}
\label{subsec:exploring_encoder}

\begin{figure}[t]
\begin{center}
\includegraphics[scale=0.6]{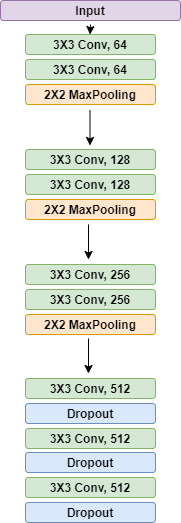}
\end{center}
   \caption{VGG-16 architecture.}
\label{fig:vgg}
\end{figure}

A modified version of VGG-16~\cite{Simonyan2015} is used as the encoder, both in FgSegNet\_v2 and the proposed method, in which the dense layers and the last convolutional block were removed, leaving the current adaptation with 4 blocks. The first 3 blocks contain 2 convolutional layers, each block followed by a MaxPooling layer. The last block holds 3 convolutional layers, each layer followed by Dropout~\cite{Cicuttin2016}. The architecture is depicted in Figure \ref{fig:vgg}.

After comparing this configuration with other models, i.e., Inception v\_3, Xception and Inception ResNet v\_2, we concluded that VGG-16 has given the best results so far, as seen in Section \ref{subsubsec:results_experiments_encoder} . This allows the extraction of the input image features with high resolution, using these features as input to the FPM.

\subsection{Feature Pooling Module}
\label{subsec:exploring_fpm}

The Feature Pooling Module receives the features from the encoder in order to correlate them at different scales, making it easier for the decoder to produce the corresponding mask.

Lim ~\cite{Lim2019} proposed a Feature Pooling Module, in which the features from a 3 X 3 convolutional layer are concatenated with the Encoder features, resulting as input to the following convolutional layer with dilation rate of 4. The features from this layer are then concatenated with the Encoder features, hence used as input to a convolutional layer with dilation rate of 8. The process repeats itself to a convolutional layer with dilation rate of 8. The output of every layer is concatenated with a pooling layer, resulting in a 5 X 64 multi-features layers. These will finally be passed through InstanceNormalization and SpatialDropout.

\begin{figure}[t]
\begin{center}
\includegraphics[scale=0.34]{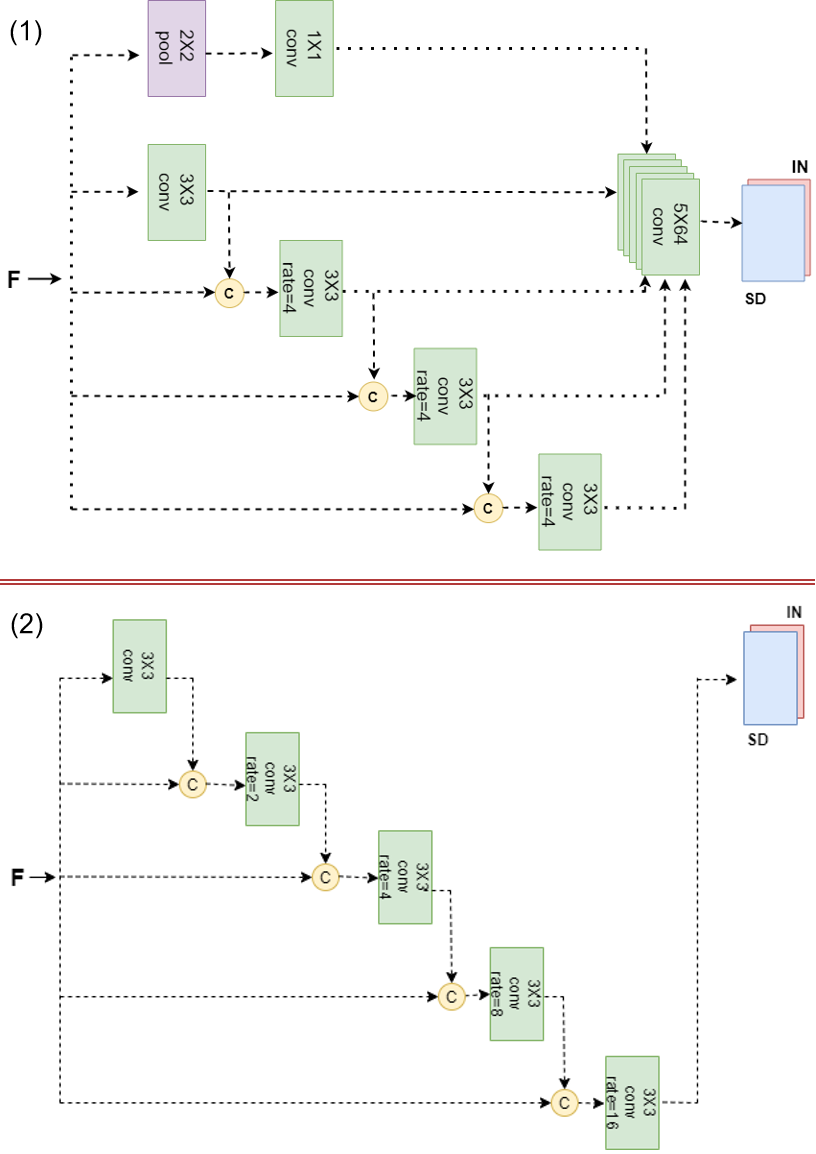}
\end{center}
   \caption{(1) Original Feature Pooling Module architecture and (2) Modified Feature Pooling Module architecture.}
\label{fig:fm}
\end{figure}

After numerous experiments and tests, an improvement to this configuration was found. A convolutional layer with dilation rate of 2 is introduced, followed by the removal of the pooling layer. The output of the last layer, i.e., convolutional with dilation rate of 16, proceeds to BatchNormalization~\cite{Ioffe2015} (providing better results than InstanceNormalization) and SpatialDropout~\cite{Cicuttin2016} (Figure \ref{fig:fm}).  

Due to the removal of the multi-features layers, SpatialDropout is used instead of Dropout. The former promoted independence between the resultant feature maps by dropping them in their entirety, correlating the adjacent pixels.

\section{\uppercase{Experiments and results}}
\label{sec:results}

\noindent When training is concluded, a test set containing the raw images and their corresponding masks is loaded, along with the trained model.

For each loaded image, a set of metrics are applied in order to evaluate the performance of the model. To complement these metrics, some visual results are also exploited. A probability mask is produced by applying a specific threshold in order to remove the low probability scores. These probabilities are translated in a Jet heatmap, i.e., when the color is red there is a high probability of containing an object and when the color is purple there is a low probability, which will be blend with the corresponding raw image, making it easier to analyse the results.


\subsection{Metrics}
\label{subsec:results_Metrics}

In order to fully understand the viability of the model, three different metrics are used throughout the evaluation.
The first one is the \emph{AUC} ( Area Under Curve )~\cite{Flach2011}, being equal to the probability that the classifier will rank a randomly chosen positive example higher than a randomly chosen negative example. This metrics allows the perception of the model detecting an object correctly, i.e., if the AUC is equal to 100\% then the model detected every object in it. Nevertheless, this metric only focus on classification successes, failing to consider classification errors and not providing information regarding the segmentation. This metric plots two parameters: \emph{TPR} (true positive rate) and \emph{FPR} (false positive rate), seen in Equations \ref{eq:auc1} and \ref{eq:auc2}.

\begin{equation}
\scalebox{.95}{
    $TPR = \frac{TP}{TP + FN}$
 }   
\label{eq:auc1}
\end{equation}

\begin{equation}
\scalebox{.95}{
    $FRP = \frac{FP}{FP + TN}$
 }   
\label{eq:auc2}
\end{equation}\\

The \emph{Accuracy} and the \emph{F-Measure} could be potential candidates, since they are highly used and usually produce reliable results, however, since the dataset is highly imbalanced when regarding the foreground and the background pixels, these two metrics are not an option, since they are highly sensitive to imbalanced data, therefore \emph{MCC} (Mathews Correlation Coefficient ) is used as an alternative~\cite{Chicco2020}, see Equation \ref{eq:mcc}. This metric ranges from $[-1,+1]$ where $+1$ represents a perfect segmentation according to the ground truth, and $-1$ the opposite.  

\begin{equation}
\scalebox{.95}{
    $MCC = \frac{ ( TP\times TN )- ( FP\times FN )}{\sqrt{( TP+FP ) ( TP+FN  ) ( TN+FN  ) ( TN+FP  )}}$
 }   
\label{eq:mcc}
\end{equation}
To complement the previous metric, \emph{mIoU} ( mean intersection over union ) is also used. IoU is the area of overlap between the predicted segmentation and the ground truth divided by the area of union between the predicted segmentation and the ground truth, see Equation \ref{eq:iou}. When applied to a binary classification, a mean is computed by taking the IoU of each class and averaging them. This metric ranges from $[0,1]$ where $1$ represents a perfect overlap with the ground truth and $0$ no overlap at all. \\
\begin{equation}
\scalebox{.95}{
$IoU = \frac{\text{Area  of  Overlap}}{\text{Area  of  Union}}$
}
\label{eq:iou}
\end{equation}

Metrics such as \emph{Recall}, \emph{Specificity}, \emph{PWC ( Percentage of wrong classifications )}, \emph{Precision} and \emph{F-Measure} will also be used in some cases, in order to allow a comparison with other state of the art methods.


\subsection{Datasets}
\label{subsec:results_datasets}

Three different datasets are used to evaluate our proposal. The first one is \emph{CityScapes}, which pretends to capture images regarding outdoor street scenes \cite{Ramos}. Every image was acquired with a camera in a moving vehicle across different months, this way different scenarios were covered, such as seasons and different cities. It has $20000$ coarse pixel-level annotations, where $5000$ of them were manually selected for dense pixel-level annotations. This way, there will be high diversity in foreground objects, background and scene layout. This dataset contains an additional challenge due to the moving camera. 

The second is \emph{CDNet2014} dataset~\cite{Wang2014}, it consists of 31 camera-captured videos with a fixed camera, containing eleven categories: Baseline, Dynamic Background, Camera Jitter, Intermittent Object Motion, Shadows, Thermal, Challenging Weather, Low Frame-Rate, Night, PTZ and Air Turbulence. The spatial resolution from the captured images ranges from 320x240 to 720x576. This dataset covers various challenging anomaly detections, both indoor and outdoor.

The last dataset is SBI2015 ( Scene Background Initialization )~\cite{Christodoulidis2015}. It contains frames taken with a fixed camera from 14 video sequences with its corresponding ground-truths. It was assembled with the objective of evaluating and comparing the results of background initialization algorithms. 

\subsection{Training details}
\label{subsec:exploring_training}

The proposed method receives a raw image and its corresponding mask, i.e., ground-truth, as input, both $512\times512$. A pre-trained VGG-16 on ImageNet is used. The first three layer blocks of the VGG-16 are frozen in order to apply transfer learning to our model.\\
Binary cross entropy loss is used as the loss function, assigning more weight to the foreground in order to address the imbalancement in the data, i.e., number of foreground pixels much higher than the background ones, and also taking in consideration when there are not any foreground pixels. The optimizer used is Adam~\cite{Kingma2015}, as it provides better results amongst other tested optimizers, i.e., RMSProp, Gradient Descent, AdaGrad~\cite{Duchi2012} and AdaDelta~\cite{Zeiler2012}, and the batch size is 4.\\
A maximum number of 80 epochs is defined, stopping the training if the validation loss does not improve after 10 epochs.
The learning rate starts at 1e-4, and is reduced after 5 epochs by a factor of 0.1 if the learning stagnates. 


\subsection{Experiments}
\label{subsec:results_experiments}

Using \cite{Lim2019} as a starting point, some ablation tests were made in CDNet2014 dataset using 25 frames for training, in order to understand the importance of the different components in the architecture and what could be improved. The metrics used throughout these evaluations are presented in \ref{subsec:results_Metrics}. 

Note that results presented in the following subsections have as only purpose the comparison of the proposed changes. Hence, these do not represent an improvement over the original FgSegNet\_v2. Only when combined do these changes provide an increase in performance. 


\subsubsection{Feature Pooling Module}
\label{subsubsec:results_experiments_fpm}

The goal is to break down the Feature Pooling Module, understanding its components and evaluate if improvements could be done. 

The first stage consisted in changing the layers with different dilation rates, these dilated convolutions are used in order to systematically aggregate multi-scale
contextual information without losing resolution~\cite{Yu2016}. The tests are (1) Removing the convolutional layer with dilation rate equal to $16$, (2) Add a convolutional layer with dilation rate equal to $2$ and (3) Add a convolutional layer with dilation rate equal to $2$ and remove a convolutional layer with dilation rate equal to $16$. 
As seen in Table \ref{fig:dilrates}, when removing the layer with dilation rate of $16$, information is going to get lost, while when adding a layer with dilation rate equal to $2$, better results are being obtained. Since adding a layer with dilation rate equal to $2$ (2) improves the overall results, it will remain for the duration of the project.\\

\begin{table*}
\centering
\caption{Changing layers with different dilation rates in the Feature Pooling Module.}
\resizebox{\textwidth}{!}{\begin{tabular}{lcccccccccccc} 
\hline
             & \multicolumn{3}{c}{ \textbf{Baseline} }                      & \multicolumn{3}{c}{ \textbf{LowFrameRate} }      & \multicolumn{3}{c}{ \textbf{BadWeather} }      & \multicolumn{3}{c}{ \textbf{CameraJitter} }      \\ 
\hline
             & \textbf{F-M.}               & \textbf{~ MCC} & \textbf{~mIoU} & \textbf{~ F-M.} & \textbf{~MCC} & \textbf{~ mIoU} & \textbf{~F-M.} & \textbf{~MCC} & \textbf{~mIoU} & \textbf{~F-M.} & \textbf{~MCC} & \textbf{~ mIoU}  \\ 
\hline
\textbf{(1)} & 0.9491                     & ~ 0.9313       & ~ 0.9317       & ~ 0.6250       & ~ 0.6870      & ~ 0.6930        & ~ 0.8916      & ~ 0.8817      & ~ 0.8927       & ~ 0.9030      & ~ 0.8912      & ~ ~0.9000        \\
\textbf{(2)} & \multicolumn{1}{l}{\textbf{0.9731}} &\textbf{~  0.9697}      &\textbf{~ 0.9721}       & \textbf{~ 0.7102}       & \textbf{~ 0.7054}      & \textbf{~ 0.7098}        & \textbf{~ 0.9172}      & \textbf{~ 0.9102}      & \textbf{~ 0.9148}       & \textbf{~ 0.9218}      & \textbf{~ 0.9199}      & \textbf{~ ~0.9205}        \\
\textbf{(3)} & \multicolumn{1}{l}{0.9528} & ~ 0.9487       & ~ 0.9510       & ~ 0.6970       & ~ 0.6890      & ~ 0.6900        & ~ 0.9034      & ~ 0.8987      & ~ 0.9019       & ~ 0.9190      & ~ 0.9076      & ~ ~0.9119        \\
\hline
\end{tabular}}
\label{fig:dilrates}
\end{table*}

The second stage is to change some of the concatenations between the layers. These tests consist in (1) Remove output from layer with dilation rate of 2 to the final concatenations, (2) Only concatenate layer with dilation rate=16 to the pooling and layer with r=1, (3) Delete every connection from pooling layer, (4) Only keep the final output of layer
with dilation rate of 16 as input to the SpatialDropout and (5) Delete every concatenation.

As seen in Table \ref{fig:concat}, when comparing the concatenations, (4) produces the best results. Hence, the other concatenation tests will be discarded.\\

\begin{table*}
\centering
\caption{Changing concatenations between layers in the Feature Pooling Module.}
\resizebox{\textwidth}{!}{\begin{tabular}{llccccccccccc} 
\hline
             & \multicolumn{3}{c}{ \textbf{Baseline} }                                                        & \multicolumn{3}{c}{ \textbf{LowFrameRate} }                                                & \multicolumn{3}{c}{ \textbf{BadWeather} }                                                  & \multicolumn{3}{c}{ \textbf{CameraJitter} }                                                  \\ 
\hline
             & \multicolumn{1}{c}{\textbf{F-M.}} & \textbf{~ MCC}               & \textbf{~mIoU}               & \textbf{~ F-M.}               & \textbf{~MCC}                & \textbf{~ mIoU}              & \textbf{~F-M.}                & \textbf{~MCC}                & \textbf{~mIoU}               & \textbf{~F-M.}                & \textbf{~MCC}                & \textbf{~ mIoU}                \\ 
\hline
\textbf{(1)} & \multicolumn{1}{c}{0.9281}       & ~ 0.9198                     & ~ 0.9260                     & ~ 0.6623                     & ~ 0.6596                     & ~ 0.6602                     & ~ 0.8916                     & ~ 0.8897                     & ~ 0.8904                     & ~ 0.9082                     & ~ 0.9047                     & ~ ~0.9058                      \\
\textbf{(2)} & 0.9528                           & ~ 0.9502                     & ~ 0.9516                     & ~ 0.6898                     & ~ 0.6885                     & ~ 0.6890                     & ~ 0.8994                     & ~ 0.8978                     & ~ 0.8983                     & ~ 0.9036                     & ~ 0.9011                     & ~ ~0.9023                      \\
\textbf{(3)} & 0.9804                           & ~ 0.9789                     & ~ 0.9799                     & ~ 0.7328                     & ~ 0.7311                     & ~ 0.7320                     & ~ 0.9207                     & ~ 0.9200                     & ~ 0.9202                     & ~ 0.9386                     & ~ 0.9368                     & ~ ~0.9374                      \\
\textbf{(4)} & \textbf{0.9824}                           & \multicolumn{1}{l}{\textbf{~ 0.9817}} & \multicolumn{1}{l}{\textbf{~ 0.9820}} & \multicolumn{1}{l}{\textbf{~ 0.7517}} & \multicolumn{1}{l}{\textbf{~ 0.7502}} & \multicolumn{1}{l}{\textbf{~ 0.7513}} & \multicolumn{1}{l}{\textbf{~ 0.9305}} & \multicolumn{1}{l}{\textbf{~ 0.9296}} & \multicolumn{1}{l}{\textbf{~ 0.9300}} & \multicolumn{1}{l}{\textbf{~ 0.9492}} & \multicolumn{1}{l}{\textbf{~ 0.9486}} & \multicolumn{1}{l}{\textbf{~ ~0.9490}}  \\
\textbf{(5)} & 0.9528                           & \multicolumn{1}{l}{~ 0.9518} & \multicolumn{1}{l}{~ 0.9520} & \multicolumn{1}{l}{~ 0.6917} & \multicolumn{1}{l}{~ 0.6909} & \multicolumn{1}{l}{~ 0.6912} & \multicolumn{1}{l}{~ 0.8923} & \multicolumn{1}{l}{~ 0.8917} & \multicolumn{1}{l}{~ 0.8920} & \multicolumn{1}{l}{~ 0.9029} & \multicolumn{1}{l}{~ 0.9010} & \multicolumn{1}{l}{~ ~0.9019}  \\
\hline
\end{tabular}}
\label{fig:concat}
\end{table*}


\subsubsection{Decoder}
\label{subsubsec:results_experiments_decoder}

The decoder in~\cite{Lim2019} uses three Convolutional layers 3X3, followed by InstanceNormalization, and multiplies the output of the first two layers with the output of the Encoder and GlobalAveragePooling (GAP).

The first stage consists in analysing the importance of the GAP, maintaining the output of the Encoder or removing it ( according to its corresponding GAP ). The tests are (1) Remove the first GAP and its corresponding Encoder's output, (2) Remove the second GAP and its corresponding Encoder's output, (3) Remove both GAP and the Encoder's output, (4) Remove first GAP but keep the corresponding Encoder's output, (5) Remove second GAP but keep the corresponding Encoder's output and (6) Remove both GAP but keep both Encoder's output. 

As seen in Table \ref{tab:GAP1}, (4) and (6) produce the worst results, decreasing the AUC, MCC and mIoU. As for the other configurations, they produce an overall increase in every metric.
When removing both GAP modules and the Encoder's outputs, option (3), the MCC and mIoU record the highest values, therefore the other configurations will be discarded.

\begin{table}
\centering
\caption{Changes in the Decoder, by changing the configuration of the GAP.}
\begin{tabular}{lllll} 
\hline
             & \multicolumn{2}{c}{\textbf{Baseline}}                                  & \multicolumn{2}{c}{\textbf{LowFrameRate}}                               \\ 
\hline
             & \multicolumn{1}{c}{\textbf{~F-M.}} & \multicolumn{1}{c}{\textbf{~MCC}} & \multicolumn{1}{c}{\textbf{~F-M.}} & \multicolumn{1}{c}{\textbf{~MCC}}  \\ 
\hline
\textbf{(1)} & 0.9527                             & 0.9502                            & 0.8010                             & 0.7927                             \\
\textbf{(2)} & 0.9683                             & 0.9650                            & 0.8438                             & 0.8412                             \\
\textbf{(3)} & \textbf{0.9856}                             & \textbf{0.9851}                            & \textbf{0.9680}                             & \textbf{0.9655}                             \\
\textbf{(4)} & 0.8716                             & 0.8678                            & 0.7617                             & 0.7600                             \\
\textbf{(5)} & 0.9628                             & 0.9611                            & 0.8017                             & 0.8000                             \\
\textbf{(6)} & 0.8847                             & 0.8826                            & 0.7729                             & 0.7711                             \\
\hline
\label{tab:GAP1}
\end{tabular}
\end{table}

After analysing the GAP, the relevance of the multiplications between the output of the first two layers in the Decoder and the output of the Encoder after applying the GAP must be considered. Therefore, additional tests were performed: (1) Remove first multiplication, (2) Remove second multiplication and (3) Remove both multiplications.

As seen in Table \ref{tab:GAP2} the results decrease when compared to (3) from the previous stage ( Table \ref{tab:GAP1} ), hence these configurations will be discarded.

\begin{table}
\centering
\caption{Changes in the Decoder, by changing the multiplications between the GAP and the dense layers.}
\begin{tabular}{lllll} 
\hline
             & \multicolumn{2}{c}{\textbf{Baseline}}                                  & \multicolumn{2}{c}{\textbf{LowFrameRate}}                               \\ 
\hline
             & \multicolumn{1}{c}{\textbf{~F-M.}} & \multicolumn{1}{c}{\textbf{~MCC}} & \multicolumn{1}{c}{\textbf{~F-M.}} & \multicolumn{1}{c}{\textbf{~MCC}}  \\ 
\hline
\textbf{(1)} & 0.8926                             & 0.8911                            & 0.8618                             & 0.8601                             \\
\textbf{(2)} & 0.9327                             & 0.9316                            & 0.9137                             & 0.9122                             \\
\textbf{(3)} & 0.9126                             & 0.9111                            & 0.9017                             & 0.8985                             \\
\hline
\label{tab:GAP2}
\end{tabular}
\end{table}


\subsubsection{Encoder}
\label{subsubsec:results_experiments_encoder}

Keeping the previous configurations of the FPM and the Decoder, three tests were made in the Encoder, changing the VGG-16 backbone to (1) Inception v3~\cite{Szegedy2016}, (2) Xception~\cite{Chollet2017} and (3) Inception ResNet v2~\cite{Szegedy2017}.

When comparing these three models with VGG-16, the latter produces better results in every metric, e.g., in the F-Measure and MCC as seen in Table \ref{tab:encoder}, while also keeping the number of parameters much lower. Therefore no changes were made when compared to the FgSegNet\_v2.


\begin{table}
\centering
\caption{Different Encoders.}
\begin{tabular}{lllll} 
\hline
             & \multicolumn{2}{c}{\textbf{Baseline}}                                  & \multicolumn{2}{c}{\textbf{LowFrameRate}}                               \\ 
\hline
             & \multicolumn{1}{c}{\textbf{~F-M.}} & \multicolumn{1}{c}{\textbf{~MCC}} & \multicolumn{1}{c}{\textbf{~F-M.}} & \multicolumn{1}{c}{\textbf{~MCC}}  \\ 
\hline
\textbf{(1)} & 0.8816                             & 0.8798                            & 0.8126                             & 0.8109                             \\
\textbf{(2)} & 0.9014                             & 0.9002                            & 0.8428                             & 0.8406                             \\
\textbf{(3)} & 0.9218                             & 0.9202                            & 0.8828                             & 0.8811                             \\
\hline
\label{tab:encoder}
\end{tabular}
\end{table}


\subsection{Final results and Comparison}
\label{subsubsec:results_experiments_final}

With the previous configurations established, some results applied to the full datasets are compared with the state of the art.

\begin{figure*}
\begin{center}
\includegraphics[scale=0.55]{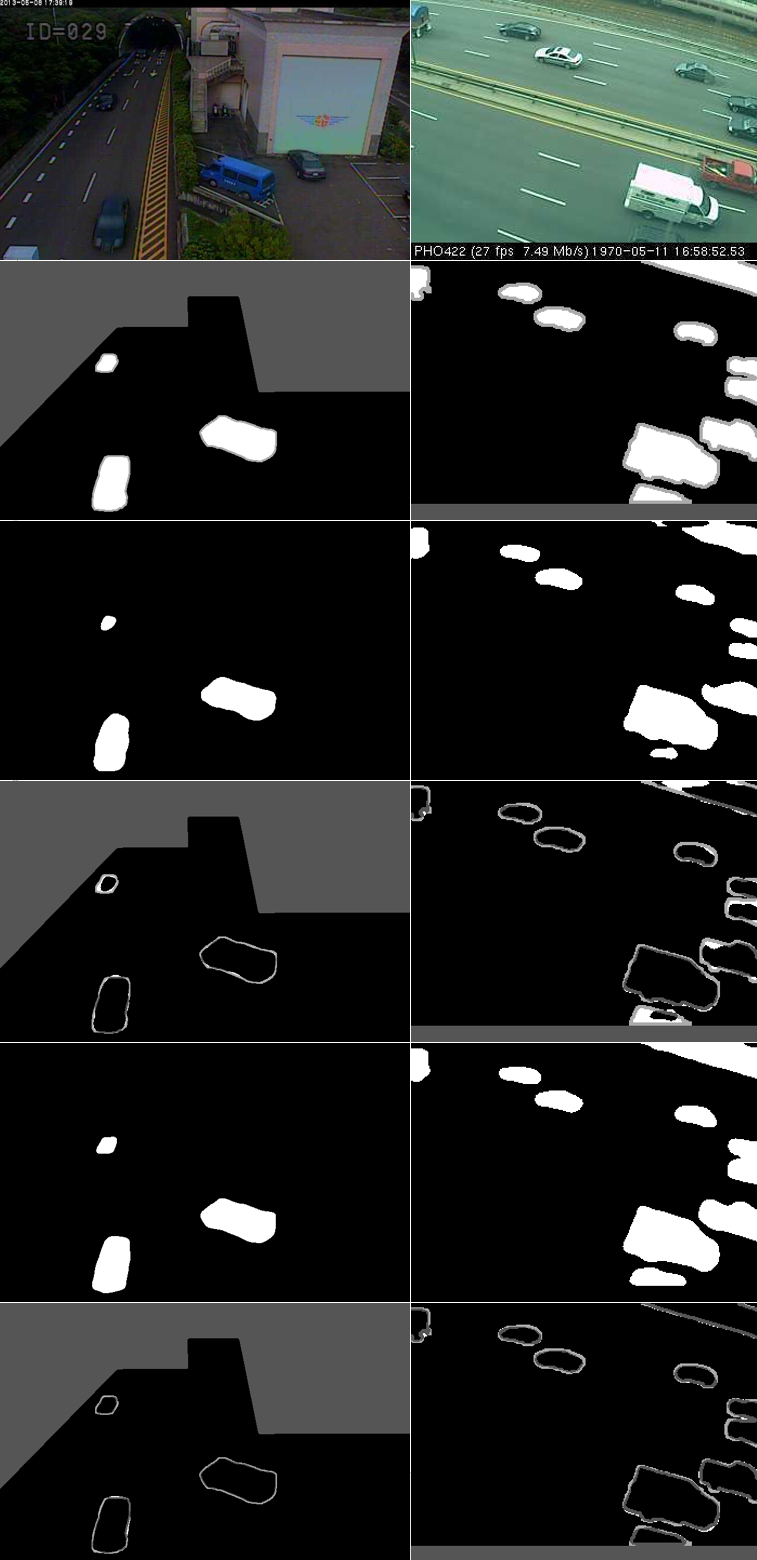}
\end{center}
   \caption{Results comparison in the LowFrameRate class. The rows represent raw images, ground-truth, FgSegNet\_v2 output masks, difference between FgSegNet\_v2 and ground truth, our proposed method output masks and difference between our proposed method and ground truth, respectively.}
\label{fig:cdnetvis}
\end{figure*}

Regarding the CDNet2014 dataset, the proposed method outperforms the top state of the art technique in this dataset, i.e., \emph{FgSegNet\_v2}, when using 200 frames for training, improving by a long margin the LowFrameRate class, going from 0.8897 to 0.9939 in the F-Measure, more details in Table \ref{tab:200_frames_comparison}. Some visual results can also be seen in Figure \ref{fig:cdnetvis}, presenting results close to the ground truth, even when dealing with LowFrameRate images.\\

\begin{table*}[!ht]
\centering
\caption{Results comparison between (1) Our proposed method and (2) FgSegnet v2 using 200 frames for training across the 11 categories.}
\begin{tabular}{llcccccccc} 
\cline{3-8}
                                       &     & \textbf{FPR} & \textbf{FNR} & \textbf{Recall} & \textbf{Precision~} & \textbf{PWC} & \textbf{F-Measure}  \\ 
\hline
\multirow{2}{*}{\textbf{Baseline~~}}   & (1) & 6e-4         & \textbf{2e-3}         & \textbf{~ 0.9979~~}      & 0.9974              & ~ 0.0129~~   & \textbf{0.9975}  \\
                                       & (2) & \textbf{4e-5}         & 3.8e-3~~     & 0.9962          & \textbf{0.9985}              & \textbf{0.0117}       & 0.9974          \\ 
\hline
\multirow{2}{*}{\textbf{Low Fr.}}      & (1) & 4.8e-3~      & \textbf{4.1e-3}       & \textbf{0.9956}          & \textbf{0.9910}              & 0.0569       & \textbf{0.9939}         \\
                                       & (2) & \textbf{8e-5}         & 9.5e-2       & 0.9044          & 0.8782              & \textbf{0.0299}       & 0.8897     \\ 
\hline
\multirow{2}{*}{\textbf{Night V.}}     & (1) & 7.4e-4       & \textbf{1.5e-2}       & \textbf{0.9848}          & 0.9785              & 0.1245       & \textbf{0.9816}                  \\
                                       & (2) & \textbf{2.2e-4}       & 3.6e-2       & 0.9637          & \textbf{0.9861}              & \textbf{0.0802}       & 0.9747                \\ 
\hline
\multirow{2}{*}{\textbf{PTZ}}          & (1) & 6e-4         & 2.3e-2       & \textbf{0.9888}          & \textbf{0.9902}              & 0.0471       & \textbf{0.9922}            \\
                                       & (2) & \textbf{4e-5}         & \textbf{2.1e-2}       & 0.9785          & 0.9834              & \textbf{0.0128}       & 0.9809        \\ 
\hline
\multirow{2}{*}{\textbf{Turbulence}}   & (1) & 3e-4         & 2.5e-2       & \textbf{0.9861}          & \textbf{0.9826}              & 0.0417       & \textbf{0.9878}               \\
                                       & (2) & \textbf{1e-4}       & \textbf{2.2e-2}       & 0.9779          & 0.9747              & \textbf{0.0232}       & 0.9762          \\ 
\hline
\multirow{2}{*}{\textbf{Bad Wea.}}     & (1) & 1.3e-4       & \textbf{1e-2}         & \textbf{0.9913}          & \textbf{0.9914}              & 0.0379       & \textbf{0.9881}                \\
                                       & (2) & \textbf{9e-5}         & 2.1e-2       & 0.9785          & 0.9911              & \textbf{0.0295}       & 0.9848                  \\ 
\hline
\multirow{2}{*}{\textbf{Dyn.Back.}}    & (1) & 3e-5         & \textbf{7.4e-3}       & \textbf{0.9958}          & \textbf{0.9959}              & 0.0067       & \textbf{0.9960}                 \\
                                       & (2) & \textbf{2e-5}       & 7.5e-3         & 0.9925          & 0.9840              & \textbf{0.0054}       & 0.9881            \\ 
\hline
\multirow{2}{*}{\textbf{Cam.Jitter}}   & (1) & 1.6e-4       & \textbf{2.5e-3}       & \textbf{0.9974}          & 0.9940              & \textbf{0.0275}       & \textbf{0.9957}                \\
                                       & (2) & \textbf{1.2e-4}       & 9.3e-3       & 0.9907          & \textbf{0.9965}              & 0.0438       & 0.9936     \\ 
\hline
\multirow{2}{*}{\textbf{Int.Obj.Mot.}} & (1) & 2.3e-4       & \textbf{7.4e-3}       & \textbf{0.9925}          & 0.9972              & 0.0672       & \textbf{0.9948}                    \\
                                       & (2) & \textbf{1.5e-4}       & 1e-2         & 0.9896          & \textbf{0.9976}              &\textbf{0.0707}       & 0.9935      \\ 
\hline
\multirow{2}{*}{\textbf{Shadow}}       & (1) & 4e-4         & 1.6e-2       & 0.9909          & 0.9942              & 0.0542       & 0.9938              \\
                                       & (2) & \textbf{1e-4}       & \textbf{5.6e-3}       & \textbf{0.9944}          & \textbf{0.9974}              & \textbf{0.0290}       & \textbf{0.9959}               \\ 
\hline
\multirow{2}{*}{\textbf{Thermal}}      & (1) & \textbf{2e-4}         & \textbf{2.7e-3}       & \textbf{0.9972}          & \textbf{0.9954}              & \textbf{0.0302}       & \textbf{0.9963}                 \\
                                       & (2) & 2.4e-4         & 8.9e-3       & 0.9911          & 0.9947              & 0.0575       & 0.9929                \\
\hline
\end{tabular}
\label{tab:200_frames_comparison}
\end{table*}

In the SBI2015 dataset, the overall F-Measure decreased from 0.9853 to 0.9447 when compared with the \emph{FgSegNet\_v2}, but increased from 0.8932 to 0.9447 when compared with the CascadeCNN,~\cite{Wang2017-2}. Nevertheless, the results still confirm a good overall evaluation on this dataset, compensating the higher results assembled in the CDNet2014 dataset. More details can be seen in Table \ref{tab:sbi}. 

\begin{table}
\centering
\caption{Results on the SBI dataset on 13 categories.}
\resizebox{\columnwidth}{!}{\begin{tabular}{lccc} 
\hline
                          & \textbf{AUC} & \textbf{F-M.} & \textbf{~ MCC}  \\ 
\hline
\textbf{Board}            & 99.84        & ~0.9734~      & ~ 0.9724        \\
\textbf{Candela}          & ~99.92~      & 0.9640        & ~ 0.9631        \\
\textbf{CAVIAR1}          & 97.25        & 0.9475        & ~ 0.9466        \\
\textbf{CAVIAR2}          & 97.51        & 0.9011        & ~ 0.9001        \\
\textbf{Cavignal}         & 99.94        & 0.9881        & ~ 0.9872        \\
\textbf{Foliage}          & 99.10        & 0.9124        & ~ 0.9115        \\
\textbf{HallAndMonitor}   & 98.49        & 0.9169        & ~ 0.9160        \\
\textbf{Highway1}         & 99.35        & 0.9593        & ~ 0.9583        \\
\textbf{Highway2}         & 99.56        & 0.9528        & ~ 0.9518        \\
\textbf{Humanbody2}       & 99.82        & 0.9579        & ~ 0.9580        \\
\textbf{IBMtest2}         & 99.42        & 0.9521        & ~ 0.9512        \\
\textbf{PeopleAndFoliage} & 99.77        & 0.9570        & ~ 0.9560        \\
\textbf{Snellen}          & 98.84        & 0.8977        & ~ 0.8967        \\

\hline
\label{tab:sbi}
\end{tabular}}
\end{table}

\begin{figure*}[!ht]
\begin{center}
\includegraphics[scale=0.72]{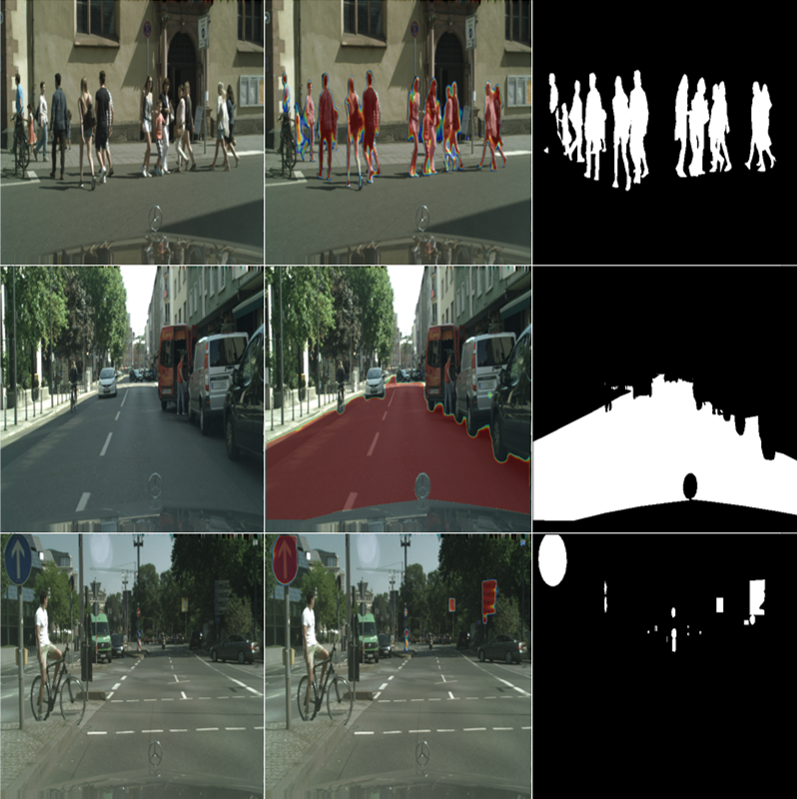}
\end{center}
   \caption{Results comparison in the CityScapes dataset. The columns represent raw images, our proposed method output mask and ground truth, respectively. The first row corresponds to Citizens class, second row to Road class and last two to Traffic Signs class.}
\label{fig:cityscapesvis}
\end{figure*}

Last, some preliminary tests ( without direct comparison to other datasets ) were made in the CityScapes dataset in order to evaluate the behaviour of the proposed method in a dataset with complex background changes, since FgSegNet\_v2 was not tested in such dataset. A test was made in three different classes (Road, Citizens and Traffic Signs). As seen in Table \ref{tab:cityscapes} and in Figure \ref{fig:cityscapesvis}, the proposed method is able to detect almost every object, confirmed by AUC metric.\\

\begin{table}
\centering
\caption{Results in the CityScapes dataset in the class Road, Citizens and Traffic Signs.}
\begin{tabular}{lccc} 
\hline
                       & \textbf{AUC}   & \textbf{MCC}    & \textbf{mIoU}    \\ 
\hline
\textbf{Road}          & 99.61 & 0.9555 & 0.9564  \\
\textbf{Citizens}      & 99.31 & 0.7552 & 0.8019  \\
\textbf{Traffic Signs} & 97.72 & 0.6618 & 0.7425  \\
\hline
\label{tab:cityscapes}
\end{tabular}
\end{table}

\newpage
\section{\uppercase{Conclusion}}
\label{sec:conclusion}

\noindent An improved FgSegNet\_v2 is proposed in the presented paper. By changing the Feature Pooling Module, i.e., deleting the pooling layer and only maintaining the output from the layer with dilation rate of 16, and the Decoder, i.e., removing the GAP modules, a more simplified and efficient approach is made, preserving the low number of needed training images feature while improving the overall results. 
It outperforms every state of the art method in the \emph{ChangeDetection2014} challenge, in particular in the \emph{LowFrameRate} images, showing a very significant improvement and also maintaining great results in the SBI and CityScapes datasets, resulting in a more generalized method than the others since no experiments have been shown when using these datasets simultaneously. As future work, we are going to focus on the \emph{CityScapes} dataset, maintaining or improving the good results in other datasets.

\section*{\uppercase{Acknowledgements}}
This work is supported by: European Structural and Investment Funds in the FEDER component, through the Operational Competitiveness and Internationalization Programme (COMPETE 2020) [Project nº 039334; Funding Reference: POCI-01-0247-FEDER-039334].

This work has been supported by national funds through FCT – Fundação para a Ciência e Tecnologia within the Project Scope: UID/CEC/00319/2019.

\bibliographystyle{apalike}
{\small
\bibliography{paper}}

\end{document}